\documentclass[final,authoryear,twocolumn,5p]{elsarticle}

\makeatletter
\def\ps@pprintTitle{}
\makeatother

\usepackage{amsmath,amssymb,amsfonts}
\usepackage{algorithmic}
\usepackage{acronym}
\usepackage{graphicx}
\usepackage{textcomp}
\usepackage{xcolor}
\usepackage[utf8]{inputenc}
\usepackage[T1]{fontenc}
\usepackage[main=english]{babel}
\usepackage{palatino}
\usepackage{enumitem}
\usepackage{bbold}
\usepackage{array}
\usepackage{url}
\usepackage{natbib}
\usepackage{subfigure}
\usepackage{booktabs}

\def\BibTeX{{\rm B\kern-.05em{\sc i\kern-.025em b}\kern-.08em
		T\kern-.1667em\lower.7ex\hbox{E}\kern-.125emX}}

\begin{document}
	
\title{Chronic Pain and Language: A Topic Modelling Approach to Personal Pain Descriptions}

\author[1,2]{Diogo A. P. Nunes}
\ead{diogo.p.nunes@inesc-id.pt}

\author[3]{Joana Ferreira Gomes}
\ead{jogomes@med.up.pt}

\author[3]{Fani Neto}
\ead{fanineto@med.up.pt}

\author[1,2]{David Martins de Matos}
\ead{david.matos@inesc-id.pt}

\address[1]{INESC-ID, Lisbon, Portugal}
\address[2]{Instituto Superior Técnico, Universidade de Lisboa, Lisbon, Portugal}
\address[3]{Biomedicina, Unidade de Biologia Experimental, Faculdade de Medicina, Universidade do Porto, Porto, Portugal}

\begin{abstract}
	Chronic pain is recognized as a major health problem, with impacts at the economic, social, and individual levels.
	Being a private and subjective experience, it is impossible to externally and impartially experience, describe, and interpret chronic pain as a purely noxious stimulus that would directly point to a causal agent and facilitate its mitigation, contrary to acute pain, the assessment of which is usually straightforward.
	Verbal communication is, thus, key to convey relevant information to health professionals that would otherwise not be accessible to external entities, namely, intrinsic qualities of the painful experience and the patient.
	Previous work has successfully applied manual linguistic analyses to the language of pain, some resulting in widely used questionnaires, in clinical settings.
	However, their fixed, lexicon-based qualities do not allow for the analysis of spontaneous verbal accounts and complex semantic and syntactic structures.
	Moreover, they cannot be easily adapted to arbitrary cultures.
	We propose a topic modelling approach to recognize complex patterns in spontaneous verbal descriptions of chronic pain, and use these patterns to quantify and qualify experiences of pain.
	Our method is implicitly subordinated to the sociocultural background of the considered patients.
	Our approach allows for the extraction of novel insights on spontaneous verbal accounts of chronic pain from the obtained topic models and latent spaces.
	We argue that our results are clinically relevant for the assessment and management of chronic pain.
\end{abstract}


\maketitle

\section{Introduction}
\label{sec:introduction}
Chronic pain is a major health problem, with impacts at the individual, social, and economic levels. 
At the individual level, it dominates a multitude of aspects of the patient's life, in most cases even extending to family and friends. 
For instance, due to immobility, muscles start to weaken, discouraging further exercise or movement.
This can lead to sleep disturbances and a vulnerable immune system, affecting the subject's psychological balance.
Adequate chronic pain assessment determines the quality of its management, which has been identified as a key issue in improving the quality of life of these patients \citep{fink2000pain}.
Since there is a manifestation of pain that is not directly visible to the outside world that is clinically relevant \citep{loeser1999pain}, a verbal interview with a patient is a key moment for pain assessment and management, in which information about the cultural, behavioural, and psychosocial dimensions of the subject in pain are conveyed, intentionally or unintentionally, in the form of expressions.
The most common expressions of chronic pain are cries, facial expressions, verbal interjections, descriptions, emotional distress, disability, and other behaviours that come as consequences of these.
The expression that is the object of study of the present work is the verbal description of the experience of pain. 
The description often-times includes valuable information about the bodily distribution of the feeling of pain, temporal patterns of activity, intensity, and others. 
Additionally, the choice of words may reflect the underlying mechanisms of the causal agent \citep{wilson2009language}, which in turn can be used to redirect the therapeutic processes.
Indeed, this forms a specific sub-language which has been studied in previous research, such as the structuring of the Grammar of Pain \citep{halliday1998grammar}, and the study of its lexical profile, resulting namely in the McGill Pain Questionnaire (MPQ), which is widely used to characterize pain from a verbal standpoint, in clinical settings \citep{katz1992measurement, sullivan1995pain}. 
However, the identified MPQ pain descriptors represent only a portion of the lexical profile of the language of pain.
Some studies have specifically stated that the fixed quality of the MPQ ultimately limits the assessment in terms of stability and predictiveness, concluding that the descriptors should be subordinated to the sociocultural, linguistic background of each patient \citep{sullivan1995pain}.
Moreover, the MPQ only accounts for the lexical profile of this sub-language, leaving out of consideration more complex structures, such as syntactic and semantic structures that appear in spontaneous accounts that can convey other informations.
Additionally, all of these studies relied on manual methods and human evaluation, rendering it expensive to replicate to other cultures and languages. 

The use of NLP techniques has the potential of overcoming all of these limitations.
The computational analysis of syntactic and semantic structures of the language of pain may yield correlations between the content of the descriptions and other relevant medical and non-medical aspects of the painful experience, allowing for a systematic and quantifiable way of characterizing pain and disease manifestations on a linguistic level.
This, in turn, has the potential to aid health professionals with the clinical assessment and management of these patients.
By modelling descriptions of pain in terms of semantic concepts, or topics, it is possible to extract information from text in an unsupervised manner without relying on the explicit analysis of syntactic structures, which is complex and fragile when dealing with spontaneous accounts.
This is fundamental in contexts such as transcriptions of natural speech, which, in general, include speech disfluencies not commonly present in written text.
This is in line with the data used in this work, which consists of transcribed verbal accounts of experiences of chronic pain, in Portuguese, guided by a fixed interview.
The interview script was designed by the authors and health professionals to focus on key cognitive aspects known to influence pain perception and description.

We propose a methodology to take advantage of topic modelling techniques to extract information, in a medically relevant way, from descriptions of chronic pain.
To the knowledge of the authors at the time of submission, this is the first proposal for computational analysis of spontaneous verbal descriptions of chronic pain experiences.
To summarize, the contributions of our work are:
\begin{enumerate}
	\item Integrate the analysis of lexical, syntactic, and semantic structures, furthering the reach of the MPQ and MPQ-based clinical analysis;
	\item Automatically account for the sociocultural subordination that the MPQ and MPQ-based questionnaires lack (and cannot easily adapt);
	\item Present, apply, and discuss the computational method to extract information from spontaneous verbal accounts of chronic pain experiences, the results of which are shown to be clinically relevant;
	\item Provide novel ways (based on our methodology) to characterize and categorize chronic pain patients, according to their spontaneous accounts, paving the way for future research to integrate NLP techniques with the clinical assessment and management of these patients.
\end{enumerate}

We start by presenting the data and its collection in a medical environment, in parallel with its intrinsic challenges.
Next, we detail and discuss a methodology to overcome such difficulties, followed by the evaluation of the obtained results. 
Finally, we present approaches to the extraction of insights of chronic pain from the obtained topic models, including the identification of clinically relevant aspects of the experience, and the definition of groups of similar patients. 

\section{Background and Related Work}
The experience of pain, dependent on its temporal pattern of activity, bodily distribution, and other aspects, is moulded by multi-domain cognitive factors, both individual and sociocultural.
Some of these, known to influence pain perception and corresponding suffering, are emotional states, beliefs, expectations, and behaviours \citep{hansen2005psychology, azevedo2012epidemiology}.
Language has been found to convey part of this information \citep{melzack1971language, wilson2009language}.
Understanding how the language of pain is used for expressing specific types of pain experiences allows us to build a linguistic model of pain descriptions.
Similar descriptions of pain might describe similar characteristics of different experiences of pain. 
Characterizing these descriptions by their semantic topics allows us to quantify the relations between different experiences in this abstract space.
Also, it may be possible to characterize specific types of pain by their associated semantic topics.
Even though not directly applied to the language of pain, NLP techniques have seen an increase in health-related applications.
An important application is the extraction of mental health features from social media texts, showing that informal language can be used to accurately classify users as having or not mental health problems \citep{yates2017depression, cohan2018smhd}.
Other works focused on extracting entities and relations, such as symptoms, also from social media texts \citep{foufi2019mining, nzali2017patients}.

Topic modelling focuses on extracting latent information in a given document from a collection.
Plus, topics can be characterized by themselves when they are attributed with ”meaning”, given the context of a problem.
Models, such as LDA \citep{blei2003latent} and NMF \citep{lee1999learning}, take a vocabulary-based representation of the collection, e.g. Bag-of-Words (BoW) or Term Frequency / Inverse Document Frequency (TFIDF), and extract topics following probabilistic and matrix factorization approaches, respectively.
Certain contexts focus on short-text, where the document length shifts from the hundreds of words to the hundreds of characters, such as data from online platforms (e.g. Twitter).
In our case, we deal with spontaneous accounts of experiences of pain, which resemble short texts.
Extracting topics from short texts presents challenges that the traditional models are not capable of efficiently overcoming, specifically the difficulty in capturing word co-occurrence information, due to noise and sparsity.

Biterm Topic Model (BTM) \citep{yan2013biterm} is a probabilistic approach that tackles sparsity in short-text document collections by virtually aggregating the whole collection into a single, long document.
Thus, instead of capturing word co-occurrence at the document level, this information is captured at the collection level, so that it describes a generative process of word co-occurrence patterns, denominated biterms, instead of documents.

Latent Feature LDA (LF-LDA) \citep{nguyen2015improving}, in contrast, incorporates latent features (semantic representations) learned from external corpora in the LDA model, specifically in the topic-to-word Dirichlet multinomial component.
This allows for the external semantic information to model topic-to-word generation when there is lack of information in the documents themselves.

Generalized Pólya Urn Dirichlet Multinomial Mixture (GPU-DMM) \citep{li2016topic} follows a similar approach, but instead of directly incorporating the external semantic information into the generative process, during the estimation step, promotes semantically related words to be assigned to the same topic.
The generative process on top of which it is built is given by the Dirichlet Multinomial Mixture (DMM) model \citep{yin2014dirichlet} that makes the simplifying assumption that each short-text document talks about one single topic, so that all words in a document are sampled independently from the same topic multinomial distribution, which in turn is sampled from the topic Dirichlet mixture of the collection.

CluWords \citep{viegas2019cluwords} exploits external semantic information by replacing each term in a document BoW representation by a meta-word, the CluWord, which represents the cluster of syntactically and semantically similar words, as determined by a pre-trained word-embedding model.
This enhanced TFIDF/CluWord representation matrix is then submitted to factorization as in the traditional NMF approach.
The main limitation of this model lies in the suitability of external resources to the problem domain.

SeaNMF \citep{shi2018seaNMF} overcomes the problems associated with short-text collections by capturing beforehand word and context vectors for the whole vocabulary and explicitly using them instead of the TFIDF representation.
This is shown to capture relevant term-context correlation that otherwise would not be fully taken advantage of by the NMF approach.
However, the limitation of this model lies in the corpus availability and vocabulary diversity.

LDA-like models under-perform in short-text contexts, when compared with NMF-like models with respect to the Pointwise Mutual Information topic coherence metric \citep{chen2019experimental}, due to the sparsity and noise of short texts, the instability of stochastic Gibbs sampling when there is insufficient term co-occurrence information, and the fact that NMF can operate in matrix representations of collections which might encode term discriminative information, such as TFIDF.
For these reasons, we are interested on short-text topic modelling following non-probabilistic approaches, specifically those based on NMF.

\section{Data Acquisition}

Our data is the result of a joint collection project with the Faculty of Medicine of Univ. of Porto.
It took place at the public hospital of São João, Porto, Portugal (CHUSJ), from Oct. 2019 to Oct. 2020. 
The dataset contains verbal descriptions of chronic pain experiences, from recorded interviews.
The patients are adults ($\ge$~18 years old), of either sex, diagnosed with osteoarthritis, rheumatoid arthritis, or spondyloarthritis, and with symptoms of chronic pain.
The dataset collection project was approved by the Ethics Committee of CHUSJ.
Data confidentiality is explicitly protected: all recordings are anonymous, and results are always presented without individual references. 

\subsection{Target information}

An interview composed of 7 questions was designed with the aim of obtaining a natural description of the patient's pain experience, in their own words, whilst directing it towards the cognitive topics relevant for pain assessment.
The script, validated by multiple health professionals included in the collection process, is as follows:

\begin{enumerate}
	\item \textit{Onde localiza a sua dor?} Where does it hurt?
	\item \textit{Como descreveria a sua dor? Como a sente/que sensações provoca?} How would you describe your pain? How do you feel it/which sensations does it cause?
	\item \textit{Como tem evoluído a intensidade da dor no último mês?} How has pain intensity evolved in the past month?
	\item \textit{Como considera que a dor tem afetado o seu dia-a-dia, nomeadamente na sua atividade física, profissional e social, e o seu estado emocional?} How would you consider pain to affect your day-to-day, namely, your physical, professional, and social activities and your emotional state?
	\item \textit{Qual considera ser a origem da sua dor?} What do you believe to be the cause of your pain?
	\item \textit{Como considera que tem evoluído a sua dor, tendo em conta o tratamento (atual) aplicado?} How would you say your pain has evolved, considering the current treatment?
	\item \textit{Como acha que irá evoluir a sua dor nos próximos meses?} How do you expect your pain to develop in the coming months?
\end{enumerate}

\subsection{Acquisition setting, challenges, and pre-processing}

\begin{figure}[hbtp]
	\centering
	\includegraphics[width=3.1in]{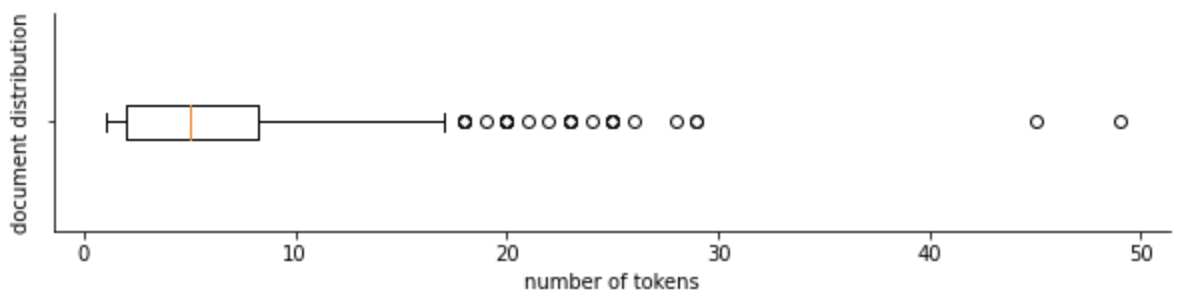}
	\caption{Document length distribution.}
	\label{fig:len-dist}
\end{figure}

Interviews with 94 patients were conducted after the pre-scheduled medical consultation with each patient, using a smartphone as a recording device.
Due to the setting of the collection, in a clinical context, there are a number of challenges: some patients showed difficulties in verbally expressing themselves; verbal accounts carry speech disfluencies, such as repetition and correction; and the medical office is not free of random environment noises.
The data pre-processing step aims at reducing the impacts of these challenges.
Interview recordings were submitted to a semi-automatic diarization process, separating patient audio segments from interview questions.
The patient audio segments were manually transcribed, leaving out speech disfluencies.
Even though the number of patients is comparable to that of other studies \citep{carlson2020characterizing, lascaratou2007language}, it is still an important limitation.
To overcome this limitation, we augmented the collection by fragmenting each patient document into the corresponding 7 answers to the interviewer.
In this setting, we are considering the answers to each question to be semantically independent.
Note that patients are still fully represented by their corresponding 7 documents.
As a result, the dataset contains 616 documents, 526 unique tokens of a total of 3872, and an average of 6.3 tokens/document.  Fig.~\ref{fig:len-dist} shows the document length distribution.
We standardize the text through lemmatization with STRING \citep{mamede2012string}, and remove stop-words as defined by the Natural Language Toolkit (NLTK) \citep{loper2002nltk} for Portuguese, with additional words empirically found not to be semantically meaningful for our task, such as the frequent formal address "senhor doutor" (doctor).
Empty documents were removed.
Even though speech disfluencies were explicitly removed, texts resulting from spontaneous speech can still be syntactically incoherent and their linguistic analysis cannot be based on the explicit modelling of syntactic structures.

\section{Topic Modelling} \label{sec:eval}

The models we apply are LDA, NMF, SeaNMF, and CluWords with pre-trained word-embeddings of both FastText \citep{mikolov2017advances} and BERT \citep{devlin2018bert}.
LDA and NMF are baselines, and are expected to under-perform in the setting of short-texts.
The others are applied in order to understand the gain in the two different types of approaches, in this case, internal and external semantic vocabulary representations, respectively.
We define $k = $ 12 as the ideal number of topics to extract from the collection, based on the design of the interview, which includes at least 7 different aspects of the experience, and because it is known that a few patients extended their descriptions beyond these aspects.
Previous experiments empirically validated this decision.

Each topic model defines a $k$-dimensional latent space over the collection of fragmented descriptions.
We evaluate the quality of each of these spaces with two metrics, regarding the model interpretability and the spatial document distribution in each topic space.

A topic is a distribution of weights over the vocabulary of the collection.
By observing the most weighted words of each topic, we can obtain an idea of the concept(s) it may cover.
Considering the various aspects of the experience of pain as dimensions on which to evaluate each patient, we expect the extracted topics to represent, in total or in part, a subset of these dimensions (as given by the interview).
A topic that does not share a single word with the remaining topics may define a concrete, modular concept, which allows for an independent evaluation of the projected population on that dimension. 
Thus, we evaluate topic modularity by determining the number of words shared between topics extracted by any one model (given the top-10 words of each topic). 
By this metric, we are most interested in the topic model which extracts topics with most unique words.
However, the discussion around this metric must take into consideration the actual words of each topic, because, according to it, a modular topic is not guaranteed to be semantically relevant.


How the projected fragments are organized in the latent semantic space reveals intrinsic qualities of that space.
Clustering and observing the Silhouette Coefficient \citep{silhouettescoreref} of each sample and cluster are two relevant measures of the spacial distribution of the fragmented descriptions.
We look at the Silhouette Coefficient of each sample in each latent space, $ s = (b-a) / \max(a,b)$, where $a$ is the mean distance between a sample and all other samples in the same cluster, and $b$ is the mean distance between that sample and all other samples in the next nearest cluster.
This provides a measure of the quality of sample distribution across clusters, and how adequate each sample is to the assigned cluster.
The mean silhouette score of a given topic model is obtained by averaging the silhouette scores of all samples in the topic space defined by that model.

In our setup (fragmented short-text documents), the typical document projection is composed of a highly weighted dimension and the remaining with infinitesimal values, i.e., only one concept is being discussed.
Thus, we expect to obtain the best clustering for a model with the number of clusters close to the number of extracted topics, in this case, $c = k =$ 12.
\section{Results}

Tables ~\ref{tab:topis-lda-en} through \ref{tab:topis-clu-fast-en} show, for each model, the top-10 words of each topic.
Starting with the baseline models, we conclude that the top words defining the NMF topics allow for a slightly easier interpretation than those of LDA.
Nevertheless, both topic models are still hard to interpret.
CluWords (FastText) topics are easier to interpret: some topics seem to relate to concrete concepts, such as pain location, intensity, and treatment.
SeaNMF, on the other hand, seems to be overfit to the collection, since the words describing each topic reveal more word co-occurrence rather than conceptual information about the latent space, rendering it very hard to interpret.

\begin{figure}[htpb]
	\centering
	\includegraphics[width=3.1in]{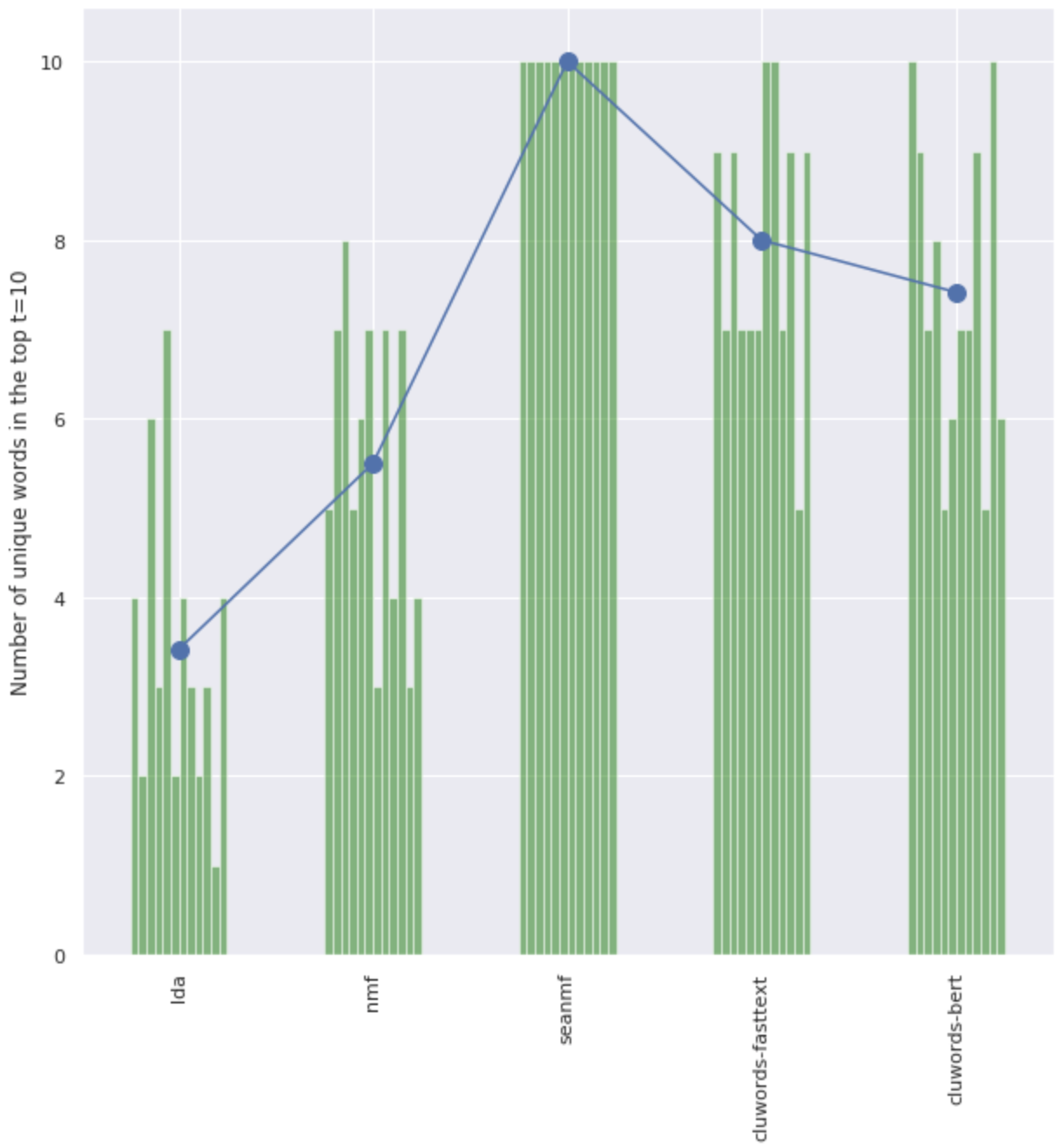}
	\caption{Modularity of each model’s extracted topic, presented in tables \ref{tab:topis-lda-en} through \ref{tab:topis-clu-fast-en}. Mean score is shown in blue.}
	\label{fig:topic-modularity}
\end{figure}

Fig.~\ref{fig:topic-modularity} shows, for each model, the topic modularity score for each topic, as well as the average modularity per model (Section~\ref{sec:eval}).
These results mirror the empirical conclusions taken from topic interpretation: the models which are hardest to interpret have the lowest scores in terms of topic modularity.
However, as observed with SeaNMF and noted in the previous section, topic modularity must never be taken into account by itself.

\begin{figure*}[htpb]
	\centering
	\includegraphics[width=6in]{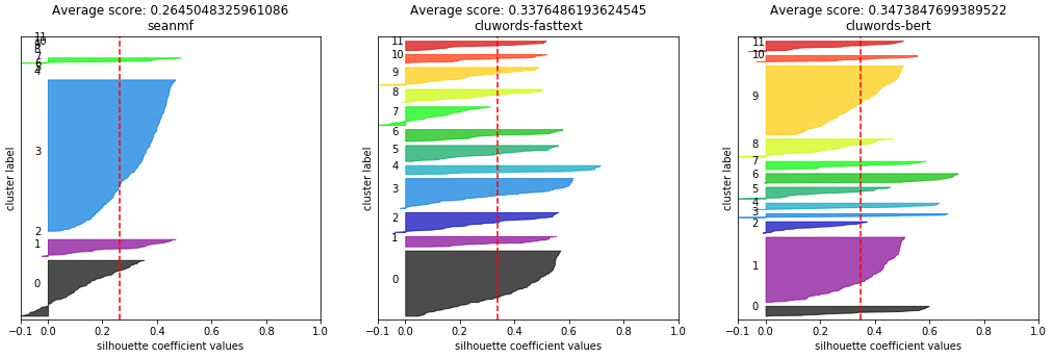}
	\caption{Silhouette scores of each sample in the topic space defined by each model.}
	\label{fig:silh}
\end{figure*}

Fig.~\ref{fig:silh} shows the Silhouette Coefficient scores for each sample for each model, for a number of clusters equal to the number of topics, $c = k = 12$. 
Baseline models were omitted for ease of discussion.
Observing the silhouette score for each sample, rather than the mean value per model, gives us a clear understanding of sample distribution in space and in relation to the assigned clusters, so that we can better assess how well the projected fragments are grouped in the topic space.
As expected from previous results and observations, the majority of the projected fragments in the SeaNMF topic space are indistinguishable from one another, and there is a poor distribution of samples per cluster, suggesting that these projection dimensions are not as semantically interesting as those of the CluWords models.
For the CluWords models, we observe that there is a better sample distribution per cluster in the CluWords (FastText) topic space, in general with better silhouettes per cluster, albeit with a marginally inferior mean Silhouette Coefficient score.

\section{Discussion}

Following the previous observations, we discard the baseline LDA and SeaNMF topic spaces.
The remaining models are all based on the same NMF model implementation and parameters, albeit on top of slightly different vocabulary-based representations of the fragment collection, obtained with external FastText and BERT pre-trained word embeddings, respectively for CluWords (FastText) and CluWords (BERT).
Results show that even though the vocabulary is very poor in terms of diversity, there are clear gains in using external semantic relations suitable to our domain, as demonstrated by the results of the CluWords models in relation to the baselines.
We conclude that the topic space given by CluWords with FastText is the most adequate for our task.

\subsection{Topic labelling}
The topmost weighted words of each topic are the ones most commonly used in similar contexts by the patients, and, thus, relate to some concept being discussed in those contexts.
This concept is not necessarily concrete, because it may cover multiple sub-concepts, nor comprehensive, because the words representing the topic may not cover the full meaning of the associated concept, and may be context dependent.
Therefore, assigning a label to a topic is both a difficult and biased task, which can introduce errors and limit generalizations.
Having this in mind, and because each question in the interview aims at specific aspects of the experience of pain, using only the 10 most weighted words of each topic, we interpret and associate a label reminiscent of the questions in the interview script (Tab. \ref{tab:topis-clu-fast-en}), which is in itself a bias.
Some topics are more concrete and easier to interpret: treatment, specific body locations, impacts (1, 2), time intervals, evolution, and generic body locations.
Others were associated with the concepts in the interview that are expected to prompt the use of such words: activity, actions, causes, intensity, and reflections.

In previous research, chronic pain episodes in clinical texts (reports written by health professionals when assessing patients suffering of chronic pain) have been systematically annotated, finding concepts such as date, pain location, cause, social and emotional effect, and others, to be covered in a high percentage of these clinical reports (100\%, 84\%, 90\%, 78\%, respectively), clearly evidencing the importance of these concepts for clinical assessment and management of these patients \citep{carlson2020characterizing}.
Because there is a match between relevant clinical concepts (found in the cited previous research) and the topics extracted with our presented methodology (specifically, with CluWords (FastText)), we argue that our results are intrinsically related to the experience of pain itself and are relevant for clinical assessment.

\subsection{Patient-level topic distribution}

\begin{figure}[htpb]
	\centering
	\includegraphics[width=3.2in]{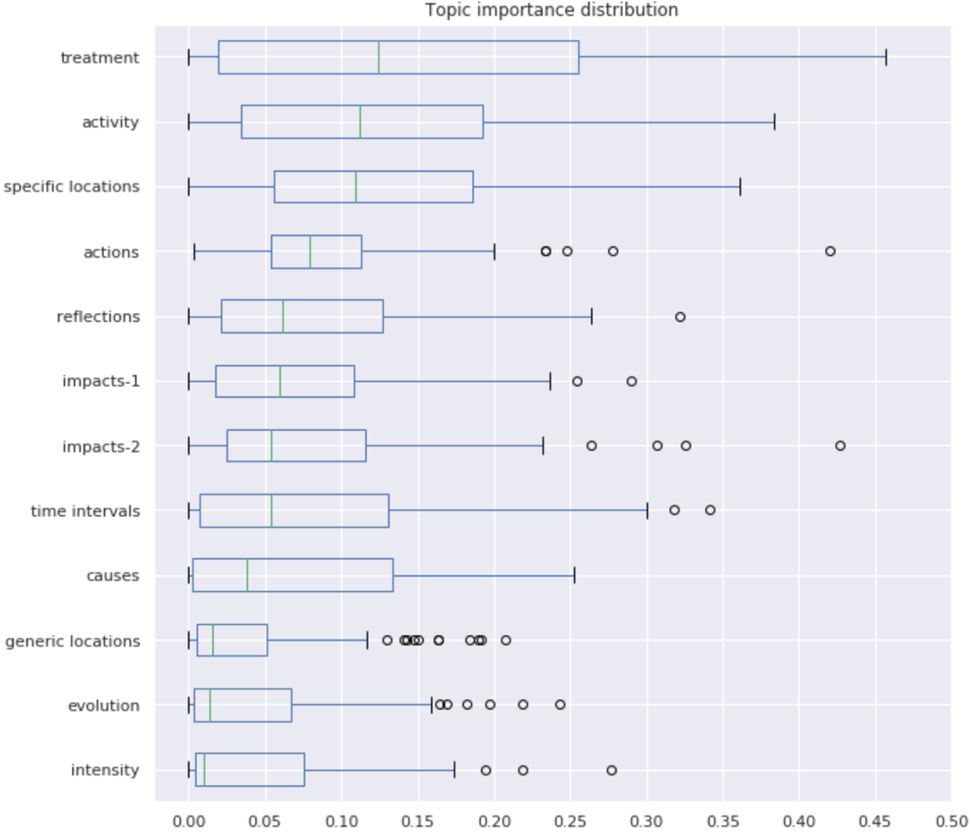}
	\caption{Distribution of topic importance over the patient-level topic projection (percentage of the total topic weight).}
	\label{fig:topic-importance-dist}
\end{figure}

In this section we demonstrate a novel way to characterize the experience of chronic pain, according to the patients considered in our dataset.
This characterization clearly highlights which dimensions of the experience of pain are more and less important for the average patient, and exactly which allow for a better distinction of patients.
Following the previous argument that the dimensions extracted with our methodology are clinically relevant, we argue that this characterization may allow for a new perspective on the clinical assessment of chronic pain patients.
If we consider each patient to be fully represented by all answers given to the corresponding interview, uniformly weighted, we obtain topic mixtures at the patient level by aggregating the corresponding fragment projections.
Thus, from the original topic projection of dimensions 616~$\times$~$k$, we obtain a patient topic distribution matrix of dimensions 94~$\times$~$k$.
We can, thus, discuss topic distributions at the patient level (Fig. \ref{fig:topic-importance-dist}).
Results show that the first three topics (treatment, activity, and specific body locations) are more relevant for patient characterization, since they display the highest variance and mean values, by large margins, with equal prompting.
Conversely, results also show that the last three topics (generic body locations, evolution, and intensity) are not relevant for patient discrimination and characterization, since around 75\% of the population assigns very low importance to those topics.
We also observe that no topic has more than 50\% importance to a patient in the collection, which adds to the notion that an experience of pain is rarely uni-dimensional.

\subsection{Patient-level similarity grouping}

\begin{figure}[htpb]
	\centering
	\includegraphics[width=3.2in]{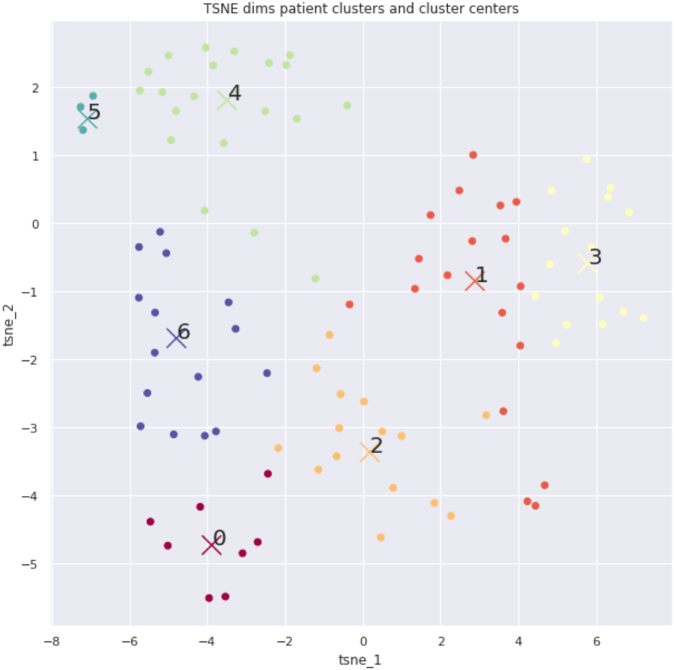}
	\caption{Projected patients on a 2D t-SNE~\citep{maaten2008visualizing} visualization, with color codes referring to the assigned cluster, for a total of 7 clusters.}
	\label{fig:patient-clusters-tsne}
\end{figure}

In this section we demonstrate a novel way to assess and manage chronic pain patients, by measuring their similarity and aggregating them according to their spontaneous descriptions.
This allows us to state that there are different groups of patients (i.e., that differ on how they experience their chronic pain), which patients belong to which group, and leverage other medical and non-medical information in each group to characterize new patients in relation to the others of the same group.
For example, when a new patient is allocated to an existing group, therapeutics and diagnosis of the other patients in that group may be informative to better clinically assess the new patient. 
Moreover, a similar characterization to that shown in the previous section can be made for each group of patients, so that each individual patient can in addition be characterized by the whole group, potentially revealing new clinical perspectives. 
Fig. \ref{fig:patient-clusters-tsne} shows a 2D visualization of patient projections and the corresponding assigned clusters and cluster centers.
Some clusters are clearly separated from the rest, whilst others have considerable dispersion.
Nevertheless, these observations show that there are patients that are similar in terms of how they describe their pain, as given by their topic mixtures, and that there are clear distinct groups.
Considering that there is a link between the description and the experience of pain, we can conclude that topic modelling over descriptions of pain may allow for the grouping of patients that share similar aspects of the experience of pain, and that this similarity is quantifiable.

\section{Conclusion and Future Work}
In this work, we presented and discussed an approach on how to take advantage of topic modelling techniques to extract interpretable information from spontaneous verbal descriptions of chronic pain.
The descriptions were obtained from recorded guided interviews, following a design shaped by the authors and specialized health professionals.
We showed the importance of the application of short-text oriented topic models in contrast to traditional approaches, and concluded on the advantage of using externally obtained semantic relations to enhance the quality of the extracted topics.
We argued on how the results are semantically relevant and intrinsically related to the experience of pain and to the clinical assessment and management of chronic pain patients.
Finally, we demonstrated novel approaches on how to extract insights from the patient level topic distributions, including the identification of the most and least relevant aspects of the experience of pain for our population, and the grouping of similar patients according to how they describe their pain.
We also reflected on how these can have an important impact in the clinical assessment and management of chronic pain patients.

To the knowledge of the authors at the time of submission, this is the first proposal to computationally analyse and characterize spontaneous accounts of chronic pain experiences.
We believe that the results presented in our work pave the way for future computational analysis of the language of chronic pain, and that this will ultimately provide new insights and techniques for the clinical assessment and management of chronic pain.

Future work, on a first step, should focus on extending the collected data, so that results may be more generalizable.
This includes both the inclusion of more patients, as well as the diversity of collection contexts, since the context in which an experience is being described highly influences the language used and the focused topics.
With larger corpora, it may then be possible to integrate the proposed analysis with data from other domains.
As an example, the proposed similarity system may be used together with diagnosis information, so that the underlying pathology causing chronic pain may be predicted for new entries (patients).
Similarly, a therapeutics-suggestion system (to help guide or inform health professionals) may be built on top of the same methodology.

\section{Acknowledgements}
Diogo A.P. Nunes is supported by a scholarship granted by INESC-ID Project UNBABEL, with reference BI|2021|133. Additionally, this work was
supported by Portuguese national funds through FCT, with reference UIDB/50021/2020.

\begin{table*}[htpb!]
	\centering
	\caption{Top-10 words of each LDA topic (each column is a topic), translated from Portuguese.}
	\resizebox{\textwidth}{!}{%
	\begin{tabular}{l|l|l|l|l|l|l|l|l|l|l|l}
		arm & bad & start & improve & think & walk & bad & day & affect & raise & want & hand \\
		leg & day & movement & treatment & wait & knee & back & medication & always & can & able to & knee \\
		work & can & walk & disease & evolve & always & hurt & take & can & explain & can & foot \\
		knee & always & joint & take & see & little & time & bad & disease & rheumatoid arthritis & move & hurt \\
		walk & hold & stiffness & medication & origin & back & reach & hurt & day & move & walk & hip \\
		home & affect & hour & time & medication & professional & professional & always & normal & able to & hand & joint \\
		hard & start & body & pass & idea & emotional & height & feel & walk & origin & people & quite \\
		position & rest & constant & big & walk & foot & start & get worse & want & foot & improve & shoulder \\
		able to & time & provoke & want & trouble & work & neck & continue & people & day & arm & wrist \\
		hip & night & less & able to & year & day & able to & walk & understand & night & stop & bone \\	
	\end{tabular}}
	\label{tab:topis-lda-en}
\end{table*}
	
\begin{table*}[htpb!]
	\centering
	\caption{Top-10 words of each NMF topic (each column is a topic, translated from Portuguese).}
	\resizebox{\textwidth}{!}{%
	\begin{tabular}{l|l|l|l|l|l|l|l|l|l|l|l}
		hurt & origin & enough & think & affect & medication & hand & want & walk & get better & wait & knee \\
		day & disease & lepicortinole & back & can & take & joint & able to & bad & treatment & evolve & foot \\
		bad & bone & life & continue & work & always & shoulder & people & be & get worse & keep & hip \\
		depend & idea & professional & problem & home & continue & knee & move & see & big & continue & shoulder \\
		start & rheumatoid arthritis & from time to time & arthritis & work & time & close & arm & leg & start & bad & wrist \\
		equal & problem & work & little & professional & leave & problem & body & hurt & medicine & idea & elbow \\
		effort & rheumatism & carry & evolve & normal & improve & able to & always & get worse & result & time & joint \\
		able to & understand & arrive & equal & tomorrow & n times & foot & emotional & work & god & stable & body \\
		night & explain & time & know & physical & disappear & bad & backward & always & see & see & finger \\
		pass & minimum & relation & level & explain & suspend & force & stop & can & truth & depend & back \\
	\end{tabular}}
	\label{tab:topis-nmf-en}
\end{table*}

\begin{table*}[htpb!]
	\centering	
	\caption{Top-10 words of each SeaNMF topic (each column is a topic), translated from Portuguese.}
	\resizebox{\textwidth}{!}{%
	\begin{tabular}{l|l|l|l|l|l|l|l|l|l|l|l}
		personal & hand & light & possible & rain & assume & word & impossible & impressive & sign & husband & valongo \\
		try & foot & cope & shape & change & muscle & fix & January & arthritis & repetitive & metro & hospital \\
		get over & knee & ask & nothing & describe & analysis & front & discomfort & drag & intensity & side to side & counts \\
		professional & hurt & task & limited & fog & special & season & resort & momentary & yesterday & diminish & find \\
		level & arm & run & side & depart & intermediate & vaccine & shreds & rheumatoid arthritis & increase & parking & horrible \\
		look & neck & heavy & health & whole & conversation & full & cut & hot & next & bottom & health professional \\
		person & leg & colleague & stable & output & improvement & rise & supermarket & disabling & difference & car & send \\
		cause & joint & work & liking & nervous & outburst & ladder & touching & weak & lingering & suffering & mouth \\
		influence & seem & go out & hold & catch & practically & crisis & chop & hot weather & lunch & walk & ui \\
		rest & close & manage & wait & excuse & common & finish & secretary & point & cortisone & walk & bed \\
	\end{tabular}}
	\label{tab:topis-seanmf-en}
\end{table*}

\begin{table*}[htpb!]
	\centering
	\caption{Top-10 words of each CluWords (BERT) topic (each column is a topic), translated from Portuguese.}	
	\resizebox{\textwidth}{!}{%
	\begin{tabular}{l|l|l|l|l|l|l|l|l|l|l|l}
		realize & huge & bone & want & bad & medication & knee & wait & strong & hand & improve & origin \\
		can & common & leg & slightly & bad & take & shoulder & evolve & intense & joint & improvement & disease \\
		take & equal & chest & get over & day & day & foot & continue & lightweight & hurt & change & idea \\
		happen & full & finger & fall & walk & always & hip & hold & one & foot & change & rheumatoid arthritis\\
		show up & pick up & run & support & back & continue & elbow & raise & a lot & back & treatment & think \\
		get up & adapt & full & be & be & hurt & wrist & try & some & problem & get worse & back \\
		namely & quite & arm & frequent & want & walk & fist & think & have & close & big & minimal \\
		consider & judge & hurt & control & backward & can & catch & see & rare & main & for now & rheumatism \\
		wrap & together & walk & ask & less & start & constant & steady & slightly & horrible & god & problem \\
		influence & meanwhile & arrest & stop & try & normal & high & desire & be & fist & eat & hip \\
	\end{tabular}}
	\label{tab:topis-clu-bert-en}
\end{table*}

\begin{table*}[htpb!]
	\centering
	\caption{Top-10 words of each CluWords (FastText) topic, preceded by a label manually defined (each column is a topic), translated from Portuguese.}
	\resizebox{\textwidth}{!}{%
	\begin{tabular}{l|l|l|l|l|l|l|l|l|l|l|l}
		Activity  & Treatment    & Specific locations & Actions & Impacts (1) & Causes & Time intervals & Intensity & Impacts (2) & Reflections & Evolution & Generic locations \\
		\hline
		start & medicine & leg & can & affect & arthritis & back and forth & enough & want & understand & improve & bone \\
		stop & treatment & shoulder & pick up & cause & disease & every now and then & less & upset & understand & improvement & muscle \\
		wait & medication & knee & take & provoke & rheumatoid arthritis & suddenly & little & get tired & explain & improvement & iliac \\
		go back & methotrexate & finger & try & depend & inflammation & every time & minimal & feel like & ask & shrink & bone \\
		linger & physiotherapy & elbow & arrive & result & pericarditis & forever & bad & think & presume & raise & neck \\
		happen & rheumatic & ankle & go back & influence & rheumatism & backward & almost & forget & respond & help & cervical \\
		continue & ointment & neck & find & change & infection & homework & practically & strive & think & change & back \\
		end & rheumatism & tendon & fail & control & rheumatic & day in the morning & totally & annoy & apologize & strain & lumbar \\
		go out & cortisone & hand & lose & decrease & medication & have to do & idea & like & talk & get worse & origin \\
		hold & take & fist & help & increase & inflammatory & suddenly & bad & try & consider & aggravate & muscle \\
	\end{tabular}}
	\label{tab:topis-clu-fast-en}
\end{table*}

\bibliographystyle{apalike}
\bibliography{references}

\begin{thebibliography}{}

\bibitem[Azevedo et~al., 2012]{azevedo2012epidemiology}
Azevedo, L.~F., Costa-Pereira, A., Mendonça, L., Dias, C.~C., and
  Castro-Lopes, J.~M. (2012).
\newblock Epidemiology of chronic pain: A population-based nationwide study on
  its prevalence, characteristics and associated disability in portugal.
\newblock {\em The Journal of Pain}, 13(8):773--783.

\bibitem[Blei et~al., 2003]{blei2003latent}
Blei, D.~M., Ng, A.~Y., and Jordan, M.~I. ({2003}).
\newblock {Latent dirichlet allocation}.
\newblock {\em {Journal of Machine Learning Research}}, {3}({Jan}):{993--1022}.

\bibitem[Carlson et~al., 2020]{carlson2020characterizing}
Carlson, L.~A., Jeffery, M.~M., Fu, S., He, H., McCoy, R.~G., Wang, Y., Hooten,
  W.~M., St~Sauver, J., Liu, H., and Fan, J. (2020).
\newblock Characterizing chronic pain episodes in clinical text at two health
  care systems: Comprehensive annotation and corpus analysis.
\newblock {\em JMIR Med Inform}, 8(11).

\bibitem[Chen et~al., 2019]{chen2019experimental}
Chen, Y., Zhang, H., Liu, R., Ye, Z., and Lin, J. ({2019}).
\newblock {Experimental explorations on short text topic mining between LDA and
  NMF based Schemes}.
\newblock {\em {Knowledge-Based Systems}}, {163}:{1--13}.

\bibitem[{Cohan, Arman and Desmet, Bart and Yates, Andrew and Soldaini, Luca
  and MacAvaney, Sean and Goharian, Nazli}, 2018]{cohan2018smhd}
{Cohan, Arman and Desmet, Bart and Yates, Andrew and Soldaini, Luca and
  MacAvaney, Sean and Goharian, Nazli} ({2018}).
\newblock {SMHD : a large-scale resource for exploring online language usage
  for multiple mental health conditions}.
\newblock In {\em {Proceedings of the 27th International Conference on
  Computational Linguistics (COLING 2018)}}, pages
  {C18--1126:1485--C18--1126:1497}. {Association for Computational Linguistics
  (ACL)}.

\bibitem[Devlin et~al., 2018]{devlin2018bert}
Devlin, J., Chang, M., Lee, K., and Toutanova, K. (2018).
\newblock {BERT:} pre-training of deep bidirectional transformers for language
  understanding.
\newblock {\em CoRR}, abs/1810.04805.

\bibitem[Fink, 2000]{fink2000pain}
Fink, R. (2000).
\newblock Pain assessment: The cornerstone to optimal pain management.
\newblock {\em Baylor Univ. Medical Center Proc.}, 13(3):236--239.
\newblock PMID: 16389388.

\bibitem[Foufi et~al., 2019]{foufi2019mining}
Foufi, V., Timakum, T., Gaudet-Blavignac, C., Lovis, C., Song, M., et~al.
  (2019).
\newblock Mining of textual health information from reddit: Analysis of chronic
  diseases with extracted entities and their relations.
\newblock {\em Journal of Medical Internet Research}, 21(6):e12876.

\bibitem[Halliday, 1998]{halliday1998grammar}
Halliday, M. (1998).
\newblock On the grammar of pain.
\newblock {\em Functions of Language}, 5(1):1--32.

\bibitem[Hansen and Streltzer, 2005]{hansen2005psychology}
Hansen, G.~R. and Streltzer, J. (2005).
\newblock The psychology of pain.
\newblock {\em Emergency Medicine Clinics of North America}, 23(2):339—348.

\bibitem[Katz and Melzack, 1999]{katz1992measurement}
Katz, J. and Melzack, R. (1999).
\newblock Measurement of pain.
\newblock {\em Surgical Clinics of North America}, 79(2):231--252.

\bibitem[Lascaratou, 2007]{lascaratou2007language}
Lascaratou, C. (2007).
\newblock {\em The Language of Pain: Expression or description?}
\newblock John Benjamins.

\bibitem[Lee and Seung, 1999]{lee1999learning}
Lee, D.~D. and Seung, H.~S. ({1999}).
\newblock {Learning the parts of objects by non-negative matrix factorization}.
\newblock {\em {Nature}}, {401}({6755}):{788}.

\bibitem[Li et~al., 2016]{li2016topic}
Li, C., Wang, H., Zhang, Z., Sun, A., and Ma, Z. ({2016}).
\newblock {Topic modeling for short texts with auxiliary word embeddings}.
\newblock In {\em {Proc. of the 39th Intl. ACM SIGIR Conf. on Research and
  Development in Information Retrieval}}, pages {165--174}. {ACM}.

\bibitem[Loeser and Melzack, 1999]{loeser1999pain}
Loeser, J.~D. and Melzack, R. (1999).
\newblock Pain: an overview.
\newblock {\em The Lancet}, 353(9164):1607--1609.

\bibitem[Loper and Bird, 2002]{loper2002nltk}
Loper, E. and Bird, S. ({2002}).
\newblock {NLTK: the natural language toolkit}.
\newblock {\em {arXiv preprint cs/0205028}}.

\bibitem[Maaten and Hinton, 2008]{maaten2008visualizing}
Maaten, L. v.~d. and Hinton, G. ({2008}).
\newblock {Visualizing data using t-SNE}.
\newblock {\em {Journal of Machine Learning Research}},
  {9}({Nov}):{2579--2605}.

\bibitem[Mamede et~al., 2012]{mamede2012string}
Mamede, N.~J., Baptista, J., Diniz, C., and Cabarr{\~a}o, V. (2012).
\newblock {STRING}: An hybrid statistical and rule-based natural language
  processing chain for portuguese.
\newblock In {\em PROPOR 2012 - 10th Intl. Conf. on Computational Processing of
  Portuguese}, Demo Session.

\bibitem[Melzack and Torgerson, 1971]{melzack1971language}
Melzack, R. and Torgerson, W. ({1971}).
\newblock {On the language of pain}.
\newblock {\em {Anesthesiology}}, {34}({1}):{50--59}.

\bibitem[Mikolov et~al., 2017]{mikolov2017advances}
Mikolov, T., Grave, E., Bojanowski, P., Puhrsch, C., and Joulin, A. (2017).
\newblock Advances in pre-training distributed word representations.
\newblock {\em CoRR}, abs/1712.09405.

\bibitem[Nguyen et~al., 2015]{nguyen2015improving}
Nguyen, D.~Q., Billingsley, R., Du, L., and Johnson, M. ({2015}).
\newblock {Improving topic models with latent feature word representations}.
\newblock {\em {Trans. of the ACL}}, {3}:{299--313}.

\bibitem[Nzali et~al., 2017]{nzali2017patients}
Nzali, M. D.~T., Bringay, S., Lavergne, C., Mollevi, C., and Opitz, T. (2017).
\newblock What patients can tell us: topic analysis for social media on breast
  cancer.
\newblock {\em JMIR Medical Informatics}, 5(3):e7779.

\bibitem[Rousseeuw, 1987]{silhouettescoreref}
Rousseeuw, P.~J. (1987).
\newblock Silhouettes: A graphical aid to the interpretation and validation of
  cluster analysis.
\newblock {\em Journal of Computational and Applied Mathematics}, 20:53--65.

\bibitem[Shi et~al., 2018]{shi2018seaNMF}
Shi, T., Kang, K., Choo, J., and Reddy, C.~K. (2018).
\newblock Short-text topic modeling via non-negative matrix factorization
  enriched with local word-context correlations.
\newblock In {\em Proc. of the 2018 World Wide Web Conf.}, WWW '18, page
  1105–1114, Republic and Canton of Geneva, CHE. Intl. World Wide Web Conf.
  Steering Committee.

\bibitem[Sullivan, 1995]{sullivan1995pain}
Sullivan, M.~D. (1995).
\newblock Pain in language: From sentience to sapience.
\newblock {\em Pain Forum}, 4(1):3--14.

\bibitem[Viegas et~al., 2019]{viegas2019cluwords}
Viegas, F., Canuto, S., Gomes, C., Luiz, W., Rosa, T., Ribas, S., Rocha, L.,
  and Gon\c{c}alves, M.~A. (2019).
\newblock Cluwords: Exploiting semantic word clustering representation for
  enhanced topic modeling.
\newblock In {\em Proc. of the 12th ACM Intl. Conf. on Web Search and Data
  Mining}, WSDM '19, page 753–761, New York, NY, USA. ACM.

\bibitem[Wilson et~al., 2009]{wilson2009language}
Wilson, D., Williams, M., and Butler, D. (2009).
\newblock Language and the pain experience.
\newblock {\em Physiotherapy Research Intl.}, 14(1):56--65.

\bibitem[Yan et~al., 2013]{yan2013biterm}
Yan, X., Guo, J., Lan, Y., and Cheng, X. ({2013}).
\newblock {A biterm topic model for short texts}.
\newblock In {\em {Proc. of the 22nd Intl. Conf. on World Wide Web}}, pages
  {1445--1456}. {ACM}.

\bibitem[Yates et~al., 2017]{yates2017depression}
Yates, A., Cohan, A., and Goharian, N. (2017).
\newblock Depression and self-harm risk assessment in online forums.
\newblock In {\em Proceedings of the 2017 Conference on Empirical Methods in
  Natural Language Processing}, pages 2968--2978.

\bibitem[Yin and Wang, 2014]{yin2014dirichlet}
Yin, J. and Wang, J. ({2014}).
\newblock {A dirichlet multinomial mixture model-based approach for short text
  clustering}.
\newblock In {\em {Proc. of the 20th ACM SIGKDD Intl. Conf. on Knowledge
  Discovery and Data Mining}}, pages {233--242}. {ACM}.

\end{thebibliography}

\end{document}